# CAD-Based Robot Programming: the role of Fuzzy-PI Force Control in Unstructured Environments

Pedro Neto, Nuno Mendes, J. Norberto Pires, *Member, IEEE*, and A. Paulo Moreira, *Member, IEEE*

*Abstract*— More and more, new ways of interaction between humans and robots are desired, something that allow us to program a robot in an intuitive way, quickly, and with a high-level of abstraction from the robot language. In this paper is presented a CAD-based system that allows users with basic skills in CAD and without skills in robot programming to generate robot programs from a CAD model of a robotic cell. When the CAD model reproduces exactly the real scenario, the system presents a satisfactory performance. On the contrary, when the CAD model does not reproduce exactly the real scenario we have an unstructured environment. In order to minimize or eliminate the previously mentioned errors, it was introduced sensory feedback (force and torque sensing) in the robotic framework. By controlling the end-effector pose and specifying its relationship to the interaction/contact forces, robot programmers can ensure that the robot maneuvers in an unstructured environment, damping possible impacts and also increasing the tolerance to positioning errors from the calibration process. Fuzzy-PI reasoning was proposed as a force control technique. The effectiveness of the proposed approach was evaluated in a series of experiments, more specifically in the manipulation of plastic cups by the robot in an unstructured environment.

## I. INTRODUCTION

OVER the last few years, a variety of solutions have been proposed to deal with robot programming. One of the main objectives of research in the robotics field is related with the development of new ways of interaction between humans and robots that allow us to program a robot in an intuitive way, quickly, and with a high-level of abstraction from the robot language. However, most of the industrial robots are still programmed by the typical teaching method, through the use of the robot teach pendant. In fact, this programming method is often tedious, time-consuming and requires some technical expertise. Therefore, new and more intuitive ways for people to interact with robots are required to make robot programming easier.

In this paper is presented a Computer Aided Design (CAD)-based system to program a robot from a 3D CAD model of a robotic cell, allowing users with basic skills in CAD and without skills in robot programming to generate robot programs off-line, without stop robot production. This system presents a satisfactory performance if the environment of the robot tasks never changes, in other words, the CAD model reproduces exactly the real scenario. However, for various reasons, in some situations the above mentioned (ideal situation) is not always achievable. In certain circumstances the robot programs are generated with errors, for example, when the CAD model does not reproduce properly the real scenario (a "foreign" object is introduced in the real environment and/or the CAD model has dimensional inaccuracies). Another factor which can affect the process is the inaccuracies created in the calibration process (virtual-real environment). In the first situation (CAD model does not reproduce properly the real scenario) we can say that we are in the presence of an unstructured environment. The development of robotic platforms in unstructured environments poses a number of challenges that cannot be easily addressed by approaches developed for highly controlled environments. In an unstructured environment, robots have to autonomously and continuously acquire the information to support decision making. In order to minimize or eliminate the previously mentioned errors, it is proposed the introduction of sensory feedback (force and torque sensing) in the robotic framework.

By controlling the end-effector pose (position and orientation) and specifying its relationship to the interaction/contact forces, robot programmers can ensure that the robot maneuvers in an unstructured environment, damping possible impacts and increasing the tolerance to positioning errors from the calibration process. Also, when any contact is made between the robot tool and its surrounding environment the interaction forces are controlled properly, otherwise the arising contact forces may damage the working objects and the robot tools. As a result, the proposed approach focuses on the use of a force control technique (Fuzzy-PI reasoning) to aid the robot performing manipulation tasks in unstructured

Manuscript received February 23, 2010. This work was supported in part by the Portuguese Foundation for Science and technology (FCT) under Grant SFRH/39218/2007.

Pedro Neto is a PhD student in the Department of Mechanical Engineering (CEMUC), University of Coimbra, Coimbra, POLO II, 3030-788, Portugal (corresponding author, phone: +351 239790700; fax: +351 239790701; e-mail: pedro.neto@dem.uc.pt).

Nuno Mendes is a PhD student in the Department of Mechanical Engineering (CEMUC), University of Coimbra, Coimbra, POLO II, 3030-788, Portugal (e-mail: nuno.mendes@dem.uc.pt).

J. Norberto Pires is with the Department of Mechanical Engineering (CEMUC), University of Coimbra, Coimbra, POLO II, 3030-788, Portugal (e-mail: jnp@robotics.dem.uc.pt).

A. Paulo Moreira is with the Institute for Systems and Computer Engineering of Porto (INESC-Porto), University of Porto, Rua Dr. Roberto Frias, 4200-465, Porto, Portugal (e-mail: amoreira@fe.up.pt).

environments. In order to proceed with the process, a force/torque (F/T) sensor was attached to the robot.

Fuzzy-PI controllers provide adequate performance when they are tuned to a specific task, but if the environmental stiffness is unknown or varying significantly, performance is degraded. The ability of the adaptable Fuzzy-PI force control system is achieved by tuning the scaling factors of the fuzzy logic controller.

The proposed CAD and force control based robotic platform increases the "intelligence" of the robot in several ways. After generate a robot program from a CAD model, the force control allows to deal with unstructured environments and help to reduce the set up time required to start robot operation. The effectiveness of the proposed approach was evaluated in a series of experiments, more specifically in the manipulation of plastic cups by the robot. The robot ability to track the desired path (extracted from CAD) and to adapt/adjust to the environment was analyzed. Finally, results are discussed and some considerations about future work directions are made.

### A. CAD-Based Robot Programming: an Overview

Since over the past few years, CAD packages are becoming more powerful and accessible, CAD-based solutions related to the robot programming task have been common. Already in the 80's, CAD was seen as a technology that could help in the development of robotics [1]. Since then, a variety of research has been conducted in the field of CAD-based robot planning and programming.

Many CAD-based systems have been proposed to assist people in the robot programming process. A series of studies have been conducted using CAD as an interface between robots and humans, for example, extracting robot motion information from a CAD data exchange format (DXF) file and converting it into robot commands [2]. A review of CAD-based robot path planning for spray painting was presented in [3]. For the same type of industrial process, a CAD-guided paint gun trajectory generation system for free-form surfaces is presented in [4]. An example of a novel process that benefits from the robots and CAD versatility is the so-called incremental forming process of metal sheets. Without using any costly form, metal sheets are clamped in a rigid frame and the robot produces a given 3D contour [5].

As we have seen above, a variety of research has been done in the area of CAD-based robot planning and programming. However, none of the studies so far deals with a "global" solution for this problem. Research studies in this area have produced great results, some of them already implemented in industry, but limited to a specific industrial process (welding, painting, etc.). Even though an abundance of approaches has been presented, a cost-effective standard solution has not been established yet.

### B. Force Control for Robotic Systems: an Overview

Many robot tasks require contact with the surrounding environment of the robot, i.e., in the process of fulfilling the task the robot tool interacts physically with the working objects and surfaces. That interaction generates contact forces that should be controlled in a way to finish the task correctly, not damaging the robot tools and working objects. Those contact forces depend on the stiffness of the tool and working objects/surfaces and should be properly controlled. The option for a particular control technique depends on identifying if [6]:

1) The contact forces should be controlled to achieve task success, but it is sufficient to keep them inside some safety domain: Passive force control [6].
2) The contact forces should be controlled because they contribute directly to the success of the task: Active force control [6]–[14].

In the first case, contact forces produce an undesirable effect on the task. They are not necessary for the process to be carried out. In the second case, the contact forces are necessary to finish the task correctly, i.e., the contact forces should be controlled, making them assume some particular value or to follow a force profile.

Up to now, many kinds of robotic systems using force control strategies have been developed and successfully applied to various industrial processes such as polishing [7] and deburring [8], [9]. A large number of force control techniques (Fuzzy, PI, PID, etc.) with varying complexity have been proposed thus far [10]–[14]. Fuzzy control was first introduced and implemented in the early 1970's [15] in an attempt to design controllers for systems that are structurally difficult to model due to naturally existing nonlinearities and other modeling complexities. Fuzzy logic control appears very useful when the processes are too complex for analysis by conventional quantitative techniques. It seems clear to everyone that force control techniques have allowed to execute increasingly more complex tasks in robotics field.

## II. PROPOSED APPROACH

In order to generate robot programs it is necessary to have information about the positions and orientations that will define the robot paths. This information is extracted from a 3D CAD model that represents the real robotic cell. Moreover, to deal with unstructured environments, a force control system is implemented in the robotic framework. The real-time force and torque feedback data from the F/T sensor will be required to achieve displacement control of the robot end-effector. By analyzing the incoming data from the F/T sensor, the implemented control system decides which displacements should be applied to achieve satisfactory robot performance, making the system adaptive to the working environment as desired. The automatic end-effector adjustment is achieved by a closed loop position control of the robot end-effector provided by F/T sensor data and Fuzzy-PI reasoning. If the contact force exceeds an acceptable range, the force control system will maneuver the robotic arm until that force falls within an acceptable range, preventing any damage. In fact, when a robot moves along a planned path (in this case extracted from CAD), the unpredictable environment uncertainties might cause damages in the robot and work environment. The introduction of force sensing in the proposed robotic platform will reduce exposure to uncertainty.

### A. System Overview

The experimental setup of the proposed robotic system is the following:
1) An industrial robot Motoman HP6 equipped with the NX100 controller.
2) A personal computer running Microsoft Windows Xp.
3) A six degrees of freedom (DOF) F/T sensor from JR3, equipped with a PCI receiver and processing board installed on the computer PCI bus.
4) A suction cup gripper attached to the F/T sensor and to the robot wrist.
5) A local area network (LAN), Ethernet and TCP/IP based, used for robot-computer communication. The network is isolated from the laboratory traffic using a properly programmed high speed (100 Mbps) network switch.

The computer is running the CAD package (Autodesk Inventor) and the software application that manages the cell (Fig. 1). The application receives data from the CAD, interprets the received data and generates robot programs. Moreover, this application manages the force control system, acquires data from the F/T sensor and sends displacement commands to the robot. For this purpose, the *MotomanLib* a Data Link Library was created in our laboratory to control and manage the robot remotely via Ethernet (Fig. 2). Using an active X component named *JR3PCI* [8], the F/T sensor measures forces and torques real-time with a sample rate of 8 KHz.

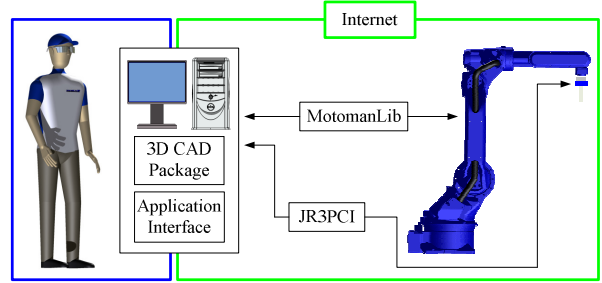

Fig. 2. Communications and system architecture.

## III. CAD-BASED INTERFACE

Once CAD technology is common throughout the industry, we propose a CAD-based system to program an industrial robot, allowing users with basic CAD skills to generate robot programs off-line. In addition, the 3D CAD package (Autodesk Inventor) that interfaces with the user is a well known CAD package, widespread in the market at a relative low-cost. This system works as a real human-robot interface where, through the CAD, the user generates programs for the real robot (Fig. 3). The methods used to extract information from the CAD (position and orientation of rigid bodies in space) and techniques to treat/convert it into robot commands are presented in [16]. The information needed to program the robot will be extracted from the CAD models through an application programming interface (API) provided by Autodesk. Experiments were conducted to evaluate the interface performance. The results showed that the CAD-based system is easy to use and within minutes an untrained user can set up the system and generate a robot program for a specific task.

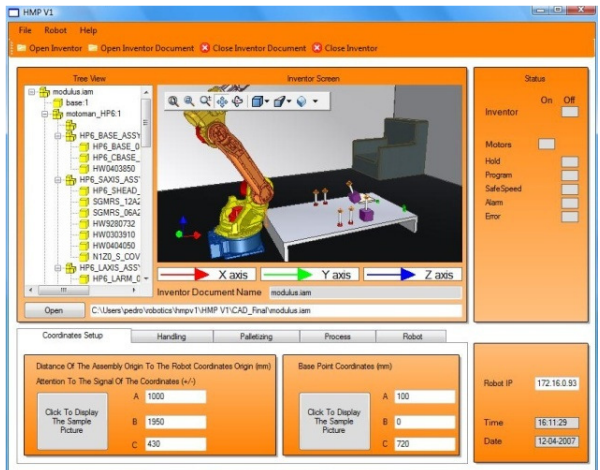

Fig. 1. Software application interface.

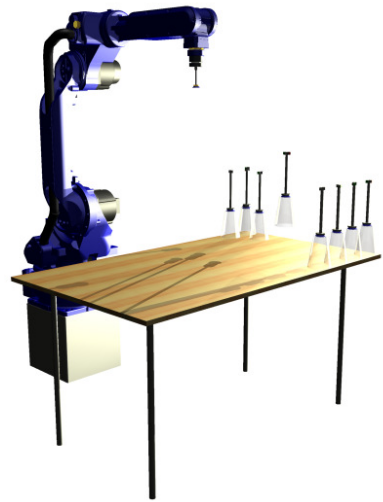

Fig. 3. CAD assembly model of the robotic material handling cell. A robot program will be generated from this model.

## IV. FORCE CONTROL

In this paper we are proposing a force control framework based on fuzzy logic to accomplish the force control problem of a robotic manipulator. The control system determines the motion of the robot (end-effector displacement) according to sensory information. Force control can be implemented using several approaches, such as PID, fuzzy, etc. However, when implementing a force controller the following conditions should be considered:
1) Simplicity. The force control law must be simple and easy/faster to compute in order to enable real-time.
2) PI-type control. If a null steady state error is achieved, a PI type force control law should be selected and implemented.
3) Implementation requirements should not include significant changes to the original control system.

Given our objective and the global system requirements (easy interpretation, implementation and at the same time flexibility) we associated two mathematical tools, the proportional integrative control (PI) and Fuzzy logic. This is a polyvalent controller with the ability to adjust to any object, regardless of its base or contact surface. Thus, a fuzzy logic controller type Mamdani based on the traditional PI controller was implemented [15].

The choice for a PI controller was due to the fact that it provides a good performance when applied in practical situations. The controller with derivative factor help to diminish the correction time, however, it is very sensitive to noise, precluding their use in our framework. Our preference for the controller type Mamdani is due to its easy implementation and the good results usually obtained, besides that do not need a rigorous mathematical model of the system. The controller is implemented only using linguistic variables to model the intuitive knowledge of the operator.

The implementation of the fuzzy controller should respect the following stages:
 --First, definition of input and output variables.
 --Second, fuzzification.
 --Third, definition of a group of rules to model the application in study (knowledge base).
 --Fourth, design of the computational unit that accesses the fuzzy rules.
 --Fifth, defuzzification.

The force control system collects inputs from the F/T sensor and processes that inputs according to the rules specified in the fuzzy logic memberships. The outputs $\Delta u$ are presented as the displacement (mm) at which the robotic arm should be moved to obtain the desired contact range of forces and torques (Fig. 4).

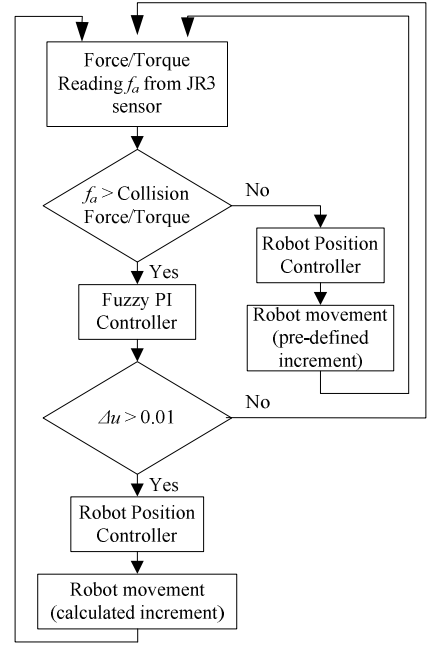

Fig. 4. Flowchart of the fuzzy-PI controller.

### A. Fuzzy Control Architecture

Fuzzy logic was conceived to apply a more human-like way of thinking in computer programming. The Fuzzy concept has been used to control the force limits of robotic arms, as the logic can handle information in a systematic way to find out precise solutions [8]. It is ideal for controlling nonlinear systems and for modeling complex systems where an inexact model exists or systems where ambiguity is common. It is also potentially very robust, maintaining good closed-loop system performance over a wide range of operating conditions. The design of a fuzzy logic controller is described by a knowledge-based system consisting of IF...THEN rules with vague predicates and a fuzzy inference mechanism.

In our system, the controller input variables are the force/torque error $e$ and change of the error $de$:

$$e_k = f_{d_k} - f_{a_k} \qquad (1)$$

$$de_k = e_k - e_{k-1} \qquad (2)$$

where $f_a$ is the actual wrench and $f_d$ is the desired wrench (set point). The controller output is the position/orientation accommodation for the robot (established in terms of robot displacements).

### B. Fuzzy-PI

From the conventional PI control algorithm:

$$u(t) = K_P \cdot e(t) + K_I \cdot \int e(t)dt \qquad (3)$$

where $u$ is the robot displacement and $K_P$ and $K_I$ are coefficients constants. Transforming (3):

$$u_k = u_{k-1} + \Delta u_k$$

$$\Delta u_k = K_P \cdot de_k + K_I \cdot e_k \qquad (4)$$

If $e$ and $de$ are fuzzy variables, (4) becomes a fuzzy control algorithm. A practical implementation of our fuzzy-PI concept is simplified in Figure 5. Finally, the center of area method was selected for defuzzify the output fuzzy set inferred by the controller (5).

$$\Delta U = \frac{\sum_{i=1}^{n} \mu_i \cdot \Delta U_i}{\sum_{i=1}^{n} \mu_i} \qquad (5)$$

where $\mu_i$ is the membership function which takes values in the interval [0, 1].

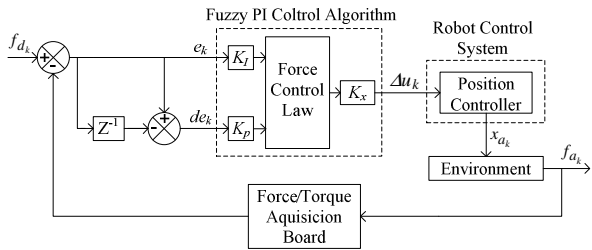

Fig. 5. Implemented fuzzy logic controller with PI function.

### C. Knowledge Base

The knowledge base of a fuzzy logic controller is composed of two components, namely, a data base and a fuzzy control rule base. Each control variable should be normalized into seven linguistic labels. The most common labels used are: positive large (PL), positive medium (PM), positive small (PS), zero (ZR), negative large (NL), negative medium (NM) and negative small (NS). The grade of each label is described by a fuzzy set. The function that relates the grade and the variable is called the membership function (Fig. 6). The well-known PI-like fuzzy rule base suggested by MacVicar-Whelan [17] is used in this paper (Table I), allowing fast working convergence without significant oscillations, and preventing overshoots and undershoots.

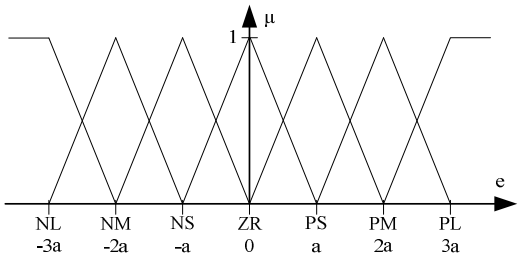

Fig. 6. Membership functions for the input variables.

TABLE I
Representation of the Rule Base

| de \ e | NL | NM | NS | ZR | PS | PM | PL |
|---|---|---|---|---|---|---|---|
| PL | nl | nm | ns | zr | pm | pl | pl |
| PM | nl | nl | nm | zr | pm | pl | pl |
| PS | nl | nl | ns | zr | ps | pl | pl |
| ZR | nl | nm | ns | zr | ps | pm | pl |
| NS | nl | nl | ns | zr | ps | pl | pl |
| NM | nl | nl | nm | zr | pm | pl | pl |
| NL | nl | nl | nm | zr | ps | pm | pl |

### D. Tuning Strategy

Fuzzy logic design is involved with two important stages: knowledge base design and tuning. However, at present there is no systematic procedure to do that. The control rules are normally extracted from practical experience, which may make the result focused in a specific application. The objective of tuning is to select the proper combination of all control parameters so that the resulting closed-loop response best meets the desired design criteria.

In order to adapt the system to various contact conditions, the scaling factors should be tuned. The controller should also be adjusted with characteristics representing the scenario to be controlled. These adjustments can be made through the scaling factors, usually applied in any PI controller, namely $K_P$, $K_I$ and $K_X$. Reference [11] proposes an adjustment where the scaling factors are dynamic and thus they have been adjusted along the task. The utilization of different tables of rules accordingly the task to be performed and the materials involved are presented in [10]. In our paper, the scaling factors are set to appropriate constant values, achieved by the method of trial and error.

## V. EXPERIMENTS, RESULTS AND DISCUSSION

The effectiveness of the proposed robotic platform was evaluated in a series of experiments, more specifically in the manipulation of plastic cups by the robot in an unstructured environment. After generating the robot program from the CAD model (Fig. 3), we force the working environment to become an unstructured environment by introducing a "foreign" object in it, in this case a hammer (Fig. 7). Without any obstruction during its operation, the robot will be moving along its pre-programmed path extracted from CAD. However, when there is contact with the "foreign" object along the way, the force control system assumes the robot control, adjusting the end-effector to the unstructured environment. The force control ensures that the contact forces and moments converge to a desired value.

We thought it was interesting to compare the performances of different force control systems. It was implement two different controllers, fuzzy-PI and PI. Force control results are presented in figure 8 (fuzzy-PI) and figure 9 (PI). The figures report the contact between the plastic cup and the hammer. Although both systems have the capability of disturbance rejection, in the PI control (Fig. 9) large overshoot is observed. Furthermore, By this reason it was established that an approach based on fuzzy-PI reasoning produce better results (an average constant force is achieved) (Fig. 8). It seems to be clear that force control improves significantly robot performance, making robots more human-like, flexible and with capacity to make decisions. As a future work, the speed of convergence to the set forces and torques should be increased.

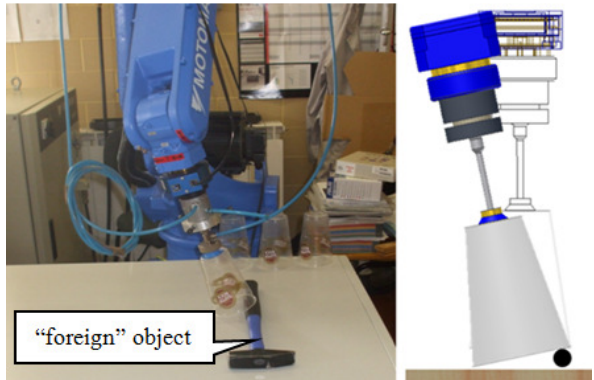

Fig. 7. Robot placing a plastic cup.

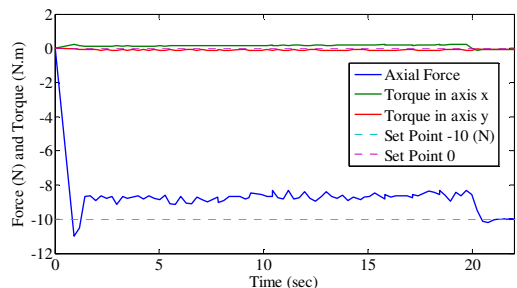

Fig. 8. Experimental results by using fuzzy-PI controller.

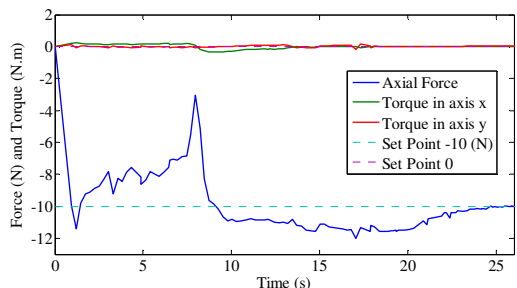

Fig. 9. Experimental results by using PI controller.

## VI. CONCLUSION

A new CAD and force based robotic platform was presented, allowing non-experts in robot programming generate robot programs from a CAD model. Owing to the implemented force control system (fuzzy-PI) the robot is able to respond to high degree of environment uncertainty (unstructured environments). The effectiveness of the proposed approach was proved through the experiments, showing that force control improves significantly robot performance, making robots more human-like, flexible and with capacity to make decisions. Also, the experiments showed that the system is intuitive and has a short learning curve, allowing users to generate robot programs in just few minutes and in an intuitive way.